\documentclass[10pt,twocolumn,letterpaper]{article}

\usepackage[pagenumbers]{cvpr}

\usepackage[dvipsnames]{xcolor}

\definecolor{cvprblue}{rgb}{0.21,0.49,0.74}
\usepackage[pagebackref,breaklinks,colorlinks,citecolor=cvprblue]{hyperref}

\usepackage{amsmath}
\usepackage{amssymb}
\usepackage{booktabs}
\usepackage{bm}

\usepackage[whole]{bxcjkjatype}
\usepackage{adjustbox}
\usepackage{multirow}
\usepackage{mathtools}
\usepackage{capt-of}
\usepackage{cuted}
\usepackage{xfrac}
\usepackage{microtype}
\usepackage{colortbl}
\usepackage{xcolor}
\usepackage{makecell}
\usepackage{pifont}
\usepackage{transparent}
\usepackage{float}
\usepackage{caption}
\usepackage{graphicx}
\usepackage{anyfontsize}

\makeatletter
\DeclareRobustCommand\onedot{\futurelet\@let@token\@onedot}
\def\@onedot{\ifx\@let@token.\else.\null\fi\xspace}

\makeatother

\renewcommand{\paragraph}{\vspace{1mm}\noindent\textbf}

\title{Introducing 3DCNN ResNets for ASD full-body kinematic assessment: a comparison with hand-crafted features}

\author{
    \href{https://orcid.org/0009-0006-3470-5836}{\includegraphics[scale=0.06]{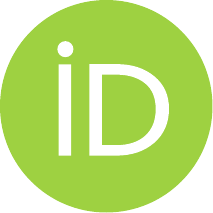}\hspace{1mm}Alberto Altozano}  \quad 
    \href{https://orcid.org/0000-0001-6326-0609}{\includegraphics[scale=0.06]{orcid.pdf}\hspace{1mm}Maria Eleonora Minissi}   \quad 
    \href{https://orcid.org/0000-0001-9207-0636}{\includegraphics[scale=0.06]{orcid.pdf}\hspace{1mm}Mariano Alcañiz}  \quad 
    \href{https://orcid.org/0000-0003-1271-2892}{\includegraphics[scale=0.06]{orcid.pdf}\hspace{1mm}Javier Marín-Morales}\\
    University Research Institute of Human-Centered Technologies\\
    Universitat Politècnica de València (UPV)\\
    {\tt\small \{aaltfer,meminiss,malcaniz,jamarmo\}@htech.upv.es \quad}
}
\usepackage{cite}
\usepackage{amsmath,amssymb,amsfonts}
\usepackage{algorithmic}
\usepackage{graphicx}
\usepackage{algorithm,algorithmic}
\usepackage{hyperref}
\usepackage{nicematrix}
\usepackage{supertabular}
\usepackage{textcomp}
\usepackage{graphicx}
\usepackage{natbib}
\usepackage{doi}

\hypersetup{
pdftitle={Introducing 3DCNN for ASD full-body kinematic assessment: a comparison with hand-crafted features},
pdfauthor={Alberto Altozazno, Maria Eleonora Minissi, Mariano Alcañiz and Javier Marín-Morales},
pdfkeywords={Autism Spectrum Discorder (ASD), Kinematic analysis, Movement data, Motor abnormalities, Feature engineering, End-to-end, Deep Learning, Spatio-temporal, Convolutional Neural Network (CNN3D), Residual Network (ResNet)},
}

\begin{document}
\maketitle

\begin{abstract}
Autism Spectrum Disorder (ASD) is characterized by challenges in social communication and restricted patterns, with motor abnormalities gaining traction for early detection. However, kinematic analysis in ASD is limited, often lacking robust validation and relying on hand-crafted features for single tasks, leading to inconsistencies across studies. End-to-end models have emerged as promising methods to overcome the need for feature engineering. Our aim is to propose a newly adapted 3DCNN ResNet from and compare it to widely used hand-crafted features for motor ASD assessment. Specifically, we developed a virtual reality environment with multiple motor tasks and trained models using both approaches. We prioritized a reliable validation framework with repeated cross-validation. Results show the proposed model achieves a maximum accuracy of 85±3\%, outperforming state-of-the-art end-to-end models with short 1-to-3 minute samples. Our comparative analysis with hand-crafted features shows feature-engineered models outperformed our end-to-end model in certain tasks. However, our end-to-end model achieved a higher mean AUC of 0.80±0.03. Additionally, statistical differences were found in model variance, with our end-to-end model providing more consistent results with less variability across all VR tasks, demonstrating domain generalization and reliability. These findings show that end-to-end models enable less variable and context-independent ASD classification without requiring domain knowledge or task specificity. However, they also recognize the effectiveness of hand-crafted features in specific task scenarios.
\footnote{This work has been submitted to Expert Systems with Applications for possible publication. Copyright may be transferred without notice, after which this version may no longer be accessible}
\end{abstract}

\section{Introduction}
\label{sec:introduction}
Autism Spectrum Disorder (ASD) is a complex neurodevelopmental condition marked by social communication difficulties and restricted, repetitive patterns of behaviors and interests\cite{DSM-5}. Detecting ASD at an early age is vital for initiating timely interventions that can significantly enhance the quality of life for affected children \cite{jl1996behavioral}. However, current diagnostic methods rely heavily on expert assessments which typically involve high costs \cite{high_costs}, highlighting the pressing need for more objective, quantifiable means of detection.

Recently, motor abnormalities have surfaced as promising early biomarkers for ASD \cite{bhat-impairments}. These abnormalities encompass a wide range of characteristics, from gross motor impairments affecting whole-body coordination and postural control to fine motor deficits affecting dexterity, handwriting, and object manipulation \cite{autism-kanner, autism-leary, autism-Mcphillips, autism-kaur, autism-lim, autism-stins, autism-dewey, autism-fleury, autism-crippa, autism-vabalas}. Additionally, stereotypical motor movements (SMMs) have also garnered significant attention for early diagnosis. SMMs refer to repetitive motions such as hand-flapping, finger-flicking, body rocking, body spinning, or head banging, that occur without a clear purpose or goal \cite{autism-peter}. Remarkably, these movements tend to emerge before the age of 3, with around 80\% of cases displaying repetitive movements by the age of 2 \cite{harris}. While the appearance rate of this symptom may not be exceptionally high, with approximately 44\% of patients reporting some form of SMM \cite{2010clinical}, it underscores the potential for quantitatively studying the motor qualities of the ASD population, in conjunction with other motor abnormalities, as a means of understanding and diagnosing ASD in young population.

Recent advancements in technology, particularly motion sensors and computer vision algorithms, have enabled the automated assessment of motor characteristics in children with ASD \cite{autism-crippa, autism-simeo,autask-zhao, autism-vabalas, autism-alcaniz, autask-zunino, autism-kojovic}. This progress has opened up new possibilities for more objective, data-driven ASD classification, aiming to reduce reliance on expert judgments and enhance the accuracy and precision of early diagnosis \cite{autisml_review}.

Currently, automatic ASD classification methods that use kinematic data often hinge on hand-crafted features meticulously designed by domain experts, with only two articles using end-to-end models \cite{autism-dl-review}. These features form the foundation for training machine learning models to distinguish between individuals with and without ASD. However, the creation and refinement of algorithms for feature extraction is a labor-intensive process, owing to their high-dimensionality, temporal dependency, sparsity and irregularity \cite{dl_4_hc}. 

In contrast, certain research domains have shifted away from hand-crafted features in favor of alternative approaches that do not require specialized expertise or extensive engineering \cite{shift_1, shift_2}. Specifically, deep learning models that operate end-to-end have gained traction as comprehensive solutions, offering superior performance over traditional techniques in various domains, action recognition being one noteworthy example \cite{arctionrecog-mldl}, which is also related to a movement classification task. However, adopting deep learning for ASD classification brings its own set of challenges, particularly related to interpretability. Despite their potential to enhance ASD classification and machine learning model development, the question remains whether these deep learning models can outperform expertly crafted features in the context of early ASD diagnosis.

Another critical aspect of the current literature on ASD motor movement analysis is the limited validation of data models \cite{autism-vabalas}. Amassing a large sample, particularly within the ASD population, poses significant challenges, resulting in studies often working with limited sample sizes. This limitation subsequently affects the size of the testing partition, with some studies forgoing validation methods which do not necessarily sufficiently control for fitting random noise in the data \cite{autism-vabalas,cv-bias}. Nevertheless, a robust validation of machine learning models, especially when dealing with health-related data, necessitates the presence of a suitably sized unseen testing partition \cite{unseen_tests}. 

Considering these aforementioned challenges, the existing literature is subject to certain limitations. Firstly, hand-crafted metrics, although developed by experts, require generating hypotheses for feature selection using human knowledge \cite{autism-dl-anibal}, potentially leading to the unintentional omission of details or the neglect of alternative metrics that could improve the accuracy of ASD classification. Additionally, across diverse experiments or methodologies, even minor variations in data yield significant variability in hand-crafted feature selection \cite{feature_selection}, making it uncertain whether they can be applied more broadly. On the other hand, using end-to-end deep learning offers a chance to create models for evaluating various methodologies by automatically generating features \cite{end-to-end-generalises}, but it comes with challenges related to explaining how the model works. Therefore, it would be beneficial for the literature to conduct a thorough performance comparison, examining results across different scenarios and assessing the trade-off between generalizability and performance. Secondly, many studies could benefit from more extensive validation procedures to predict how well a model will perform on new data accurately.

To overcome these limitations, we propose two primary strategies. Firstly, we aim to develop multiple classification models based on the body movements of children engaged in various motor tasks within a virtual reality (VR) environment. These models will utilize expert-defined metrics that have been previously employed in the literature \cite{autism-crippa, autism-vabalas, autism-forti, autism-alcaniz, autism-simeoli, autask-zhao} to predict whether each subject's sample belongs to the ASD or typically developing (TD) population. We will compare these models with a novel, fully automated model designed for ASD classification. This approach entails the use of an end-to-end deep learning model capable of automatically detecting ASD without the need for manually defined metrics. This automated approach eliminates the need for metric engineering, allowing the model to autonomously extract its own features. Similar to the hand-crafted models, we will train one end-to-end model for each VR task, enabling us to perform a comparative analysis of model performance across different scenarios.

Secondly, to ensure a fair comparison and establish the validity of the trained models, we have implemented a robust validation strategy. Our strategy involves nested subject-dependent repeated cross-validation, utilizing a dataset comprising 81 subjects (39 with ASD and 42 typically developing).  Our objective with this strategy is to ensure consistent and dependable performance that effectively generalizes to real-world, unseen data. 

In summary, the main contributions of the work are (1) introducing a newly trained 3DCNN ResNet tailored for end-to-end kinematic ASD classification, using an existing deep learning architecture from action recognition; (2) demonstrating superior performance in ASD assessment compared to previous State of the Art end-to-end models; (3) emphasizing model reliability through a dedicated focus on repeated cross-validation techniques, ensuring a reliable and accurate performance estimation; and (4) showcasing the end-to-end model's capacity for enhanced generalization across various specific motor domain datasets compared to feature-engineered models.

\section{Related Work}
\label{sec:relatedwork}

In this section, we provide an overview of previous studies specifically focusing on motor-based ASD classification with a particular focus on hand-crafted features, robust validation methods, the utilization of multiple tasks for classification, and the application of deep learning models.

Several studies have employed hand-crafted features to characterize motion patterns associated with ASD. For example, Crippa \textit{et al.} \cite{autism-crippa} employed metrics such as total movement, peak velocity, acceleration, and deceleration in their analysis. Simeo \textit{et al.} \cite{autism-simeo} used features such as average speed, average maximum and minimum speed, and acceleration, while Zhao \textit{et al.} \cite{autask-zhao} incorporated parameters like amplitude, entropy, mean, and maximum values of velocity and acceleration. However, that these studies often grapple with limitations in their validation procedures. For instance, Crippa \textit{et al.} \cite{autism-crippa} selected the feature set with the best testing performance, potentially introducing bias into their results. Simeo \textit{et al.}'s \cite{autism-simeo} cross-validation method may not be subject-dependent, which could affect the model's generalizability. Similarly, Zhao \textit{et al.} \cite{autask-zhao} explored every feature combination and reported the one with the best results, raising concerns about overfitting. These studies underscore the challenges in achieving unbiased validation in ASD classification models based on hand-crafted features.

In contrast, Vabalas \textit{et al.} \cite{autism-vabalas} prioritized robust validation processes to enhance result reliability. Their approach involved nested cross-validation along with an additional validation group, ensuring a more rigorous assessment of model performance. By testing the model on unseen data, they mitigated concerns of overfitting. Their model, utilizing support vector machines (SVMs) and feature selection, achieved a 73\% accuracy on ASD classification based on motor movement. Vabalas \textit{et al.}'s study exemplifies a commitment to dependable validation in the realm of ASD classification.

Alcañiz \textit{et al.} \cite{autism-alcaniz} pursued a distinctive approach by exploring the use of multiple VR tasks for ASD classification. Utilizing VR environments in psychological research has shown promising results in previous studies \cite{vr_review}. However, Alcañiz et al. \cite{autism-alcaniz} was the first study to combine the application of VR and ASD motion classification. Their innovative experimentation involved 24 ASD and 25 TD participants aged 4 to 7. Metrics related to the total body movement range of various body parts were extracted and used to train a SVM with recursive feature elimination (RFE). This approach resulted in an 80.29\% accuracy. Notably, their study leveraged the potential of employing a diverse set of tasks to enhance ASD classification through motion analysis while being able to test a model on multiple scenarios in order to compare task performance and model generalizability.

Deep learning models have also made a significant impact on motion-based ASD classification. Zunino \textit{et al.} \cite{autask-zunino} harnessed the power of a convolutional neural network (CNN) coupled with a long short-term memory (LSTM) network to analyze short raw videos of subjects engaged in reach and grab tasks. This innovative approach achieved a 75\% accuracy, showcasing the feasibility of deep learning techniques in motion analysis for ASD classification. In a parallel effort, Kojovic \textit{et al.} \cite{autism-kojovic} conducted an expansive study involving 169 subjects, a majority of whom had ASD. They employed deep learning techniques, including skeleton-based body tracking, and achieved an impressive 82.98\% accuracy. This study underscored the potential of deep learning in deciphering complex motion patterns linked to ASD.

In our study, we have synthesized key elements from these preceding works in motor-based ASD classification. We incorporated hand-crafted features akin to those employed by Zhao \textit{et al.}\cite{autask-zhao}, Simeo \textit{et al.} \cite{autism-simeo}, and Crippa \textit{et al.}\cite{autism-crippa} for characterizing motion. Furthermore, we adopted the strategy of Alcañiz \textit{et al.} \cite{autism-alcaniz} by using multiple VR tasks to train models based on these hand-crafted features. Additionally, we explored the potential of end-to-end deep learning models, as demonstrated by Zunino \textit{et al.} \cite{autask-zunino} and Kojovic \textit{et al.} \cite{autism-kojovic}. Our primary objective is to comprehensively assess the performance of these approaches, highlighting their performance and generalization advantages and challenges in the context of early ASD assessment. Notably, our study places a strong emphasis on robust validation, utilizing subject-dependent repeated cross-validation in every task model to ensure the reliability of our results and generalizability of our findings across different scenarios and tasks.

\section{Materials}
\label{sec:materials}

In our research, we adopted a VR approach to classify ASD and TD individuals from multiple virtual tasks, drawing inspiration from the work of Alcañiz \textit{et al.} \cite{autism-alcaniz}. The utilization of VR aimed to create a heightened sense of presence, fostering more realistic responses and facilitating the collection of organic and ecological data \cite{vr_presence}. This approach may offer advantages for ASD classification as it allows researchers to immerse subjects in controlled environments with socially-relevant and motor interactive scenarios, while maintaining a high degree of scalability and standardization.

\subsection{Participants}

In total, 81 children (42 with TD and 39 with ASD) took part in the study. Participants' ages ranged between 3 and 7 years. The group of children with ASD was composed of 32 males and 7 females, and their mean age in months was 53.14 (SD = 12.38). The group of children with TD gathered 19 males and 23 females with a mean age in months of 57.88 (SD = 11.62). The sex imbalance between groups was in line with the prevalence ratio of the disorder (4 males, every 1 female diagnosed \cite{fombonne}). Children in the ASD group had a previous ASD diagnosis made by the administration of the Autism Diagnostic Observation Schedule-2 (ADOS-2; Lord et al. \cite{lord-c-diagnostic}). Some of them were also assessed by expert neuropsychiatrics to confirm the diagnosis according to the Spanish public health system. Caregivers of ASD children were asked to provide the ASD assessment report of their children the day of the study. Moreover, participants in the ASD group underwent assessments by expert clinicians of a neurocognitive centre to ensure the absence of comorbid cognitive and language impairments, as well as anxiety, personality, and other neurodevelopmental disorders. Conversely, children in the TD group were required to have no risk or diagnosis of physical or psychological disorders, as confirmed by caregivers. Participants of both groups were Spanish and right-handed. They were drug naïve and had normal or corrected to normal vision. Additionally, caregivers of all participants provided signed consent agreements before the virtual experience commenced.


\begin{table*}
    \centering
    \begin{tabular}{|c|c|c|p{10cm}|}
    \hline
        Task Name & Abbreviature & Block & Description/Task Objective  \\ \hline
        VE  Presentation & PEAP & - & Both virtual avatars introduce themselves and familiarize the participant with the virtual environment. The participant is expected to remain still.    \\ \hline
        Introduction & I2 & - & The principal avatar asks three questions to the participant regarding their well-being, favorite game, and preferred means of transportation, which the participant is required to verbally answer. If needed, pictograms appear to pick a response by pointing at it.  \\ \hline
        Bubble Task & T2A1 & A &  Participants must interact with the virtual environment by touching and blowing up 30 descending bubbles, each with different speed levels.  \\ \hline
        Apple Task & T2A2 & A & Participants are tasked with grabbing an apple hanging from a virtual tree and placing it on the floor, repeating this action five times (top to bottom movement). \\ \hline
        Kick Task & T2A3 & A &  The principal avatar passes a ball to the participant, who is invited to kick the ball three consecutive times.  \\ \hline
        Flower Task & T2A4 & A & Participants must pick a virtual flower nearby and place it on a bench, repeating this action five times (left to right movement).  \\ \hline
        Hide \& Seek Task & T2A5 & A & The principal avatar hides in the virtual park, and the participant must indicate the avatar's hiding spot by pointing at it. This is repeated three times with different hiding spots.   \\ \hline
        Step Task & T2B1 & B & Participants are asked to imitate a sideways step movement demonstrated by the principal avatar, repeating this action five times.  \\ \hline
        Posture Task & T2B2 & B & Participants are asked to imitate a specific posture demonstrated by the principal avatar, repeating this action five times.  \\ \hline
        Highfive Task & T2B3 & B & Participants are required to virtually highfive the principal avatar five times.  \\ \hline
        Greeting Task & T2B4 & B & Participants are required to greet the principal avatar five times using virtual interaction.  \\ \hline
        Final Scene & EF & A/B & The principal avatar concludes the virtual experience by asking the participant about their favorite game in the virtual environment. If needed, pictograms appear to pick a response by pointing at it.  \\ \hline
    \end{tabular}
    \caption{Description of the proposed VR tasks.}
    \label{task}
\end{table*}

\subsection{The Experimental Setup and data collection}

Due to concerns regarding discomfort in individuals with ASD while using traditional head-mounted displays (HMDs) \cite{minissi-usability}, we opted for the CAVE Automatic Virtual Environment (CAVE) as our VR system \cite{cave-herrman, cave-use7, cave-use5}. Particularly, the extended exposure to HMDs can raise concerns for young children with severe autism, as they may experience an abrupt plunge into a virtual world that seems detached from the actual external environment \cite{Bradley_hdm}. This could also arise symptoms related to their sensory disfunction. Furthermore, it is typically not advised for young children to engage with HMDs for prolonged durations because their visual systems are still developing \cite{newbutt_hdm}.The CAVE system stands out as a favorable choice due to its capability to monitor the full-body movements of the participants, who can play freely in the room without the necessity for wearable sensors. In addition, unlike traditional HMDs, the CAVE setup eliminates problems related to cybersickness and the discomfort caused by ill-fitting HMDs, which can be particularly challenging for young individuals with ASD. The CAVE room (4 m x 4 m x 3 m) is equipped with three ultra-short lens projectors positioned in the ceiling, projecting wide 100° images at a distance of 55 centimeters. The main components of the virtual scenes were displayed on the central (3 m x 4 m) wall, while the projections on the two (4 m x 4 m) lateral surfaces further enhanced the participants' sense of being within the virtual environment.

To enable participants' interactions within the virtual environment, we employed an Azure Kinect DK, equipped with an RGB-D camera capable of capturing 640x576 pixels at 30 frames per second. The camera, in conjunction with the Azure Kinect Body Tracking SDK real-time computer vision algorithm, tracked 32 different body joints representing the user's body position. As a result, we were able to create a dynamic silhouette of the participant, accurately mirroring their movements into the projected virtual environment. Additionally, the data obtained from the computer vision algorithm was recorded as temporal sequences of the 32 body parts and stored in a text file, with each line containing the 3D position of each joint, a timestamp, and a unique identifier for each detected body in the scene.

The selection of the Azure Kinect DK and the Azure Kinect Body Tracking SDK was based on their affordability, ease of implementation, and alignment with our scientific requirements. While previous studies have noted limitations of the device and its associated computer vision algorithm in challenging scenarios such as suboptimal lighting, occlusion, and significant distances from the sensor \cite{doi:10.3390/app11125756, ieee.9480177}, our experimental setup aimed to address and mitigate these concerns. Specifically, our virtual interaction design was crafted to minimize occlusions, participants were instructed to maintain an optimal sensor distance (1 to 2 meters), and controlled diffused lighting was employed throughout the experiment.Under these controlled conditions, recent research by Romeo et al. \cite{ieee.9480177} reported minimal errors in static pose estimation, ranging from 3 to 10 mm on root mean square (RMSE) error. Additionally, Amadeus et al. \cite{pmid.33119606} demonstrated comparable joint estimation accuracy to gold standard gait tracking devices, further validating our choice. Furthermore, Bertram et al. \cite{doi:10.1371/journal.pone.0279697} reported a maximum joint RMSE tracking error of 89mm during standard clinical motor function tasks akin to ours. While acknowledging that the Azure Kinect may not be optimally suited for tasks requiring precise fine motor control such as handwriting, its performance is described as moderate to excellent in studies focusing on gross motor tasks akin to ours.

\subsection{The Virtual Experience}

The virtual environment, developed using Unity, simulated a playpark within an urban setting. It featured two virtual avatars: a child-like principal avatar, fostering social interaction and offering guidance on a series of engaging games and tasks, and a virtual therapist avatar, an adult figure that stepped in to assist participants whenever their interactions deviated from the expected behavior. The therapist avatar was a source of reassurance, providing helpful explanations to aid participants in completing the tasks.

In Table \ref{task}, we present the 12 tasks included in the virtual experience, along with their abbreviations, block assignments, and objectives. Specifically, block A refers to tasks that involve interacting with the virtual environment, while block B focuses on gesture imitation. Tasks were designed in collaboration with expert clinicians working everyday with ASD children in therapeutic settings. Participants were given 45 seconds to make progress in each task objective, with three consecutive failures leading to task termination and to the initialization of the next task.

At the beginning of the study, participants familiarized with the CAVE system and the experimenter. They then chose to interact virtually through either a male or female silhouette projected by the Azure Kinect DK in the virtual environment, with the experimenter assisting them in recognizing themselves within the silhouette. During the VR experience, to ensure that every participant had a diverse and unbiased experience, the order of blocks and tasks within each block was randomized. However, the presentation and introduction always occurred at the beginning of the experience, while the final scene was reserved for the end, regardless of the block order. Lastly, during the VR experience, the experimenter remained in a specified area of the CAVE, outside the Azure Kinect DK's tracking zone. 

\section{Methods}

\begin{figure*}
    \centering
    \includegraphics[width = 170mm, height = 105mm]{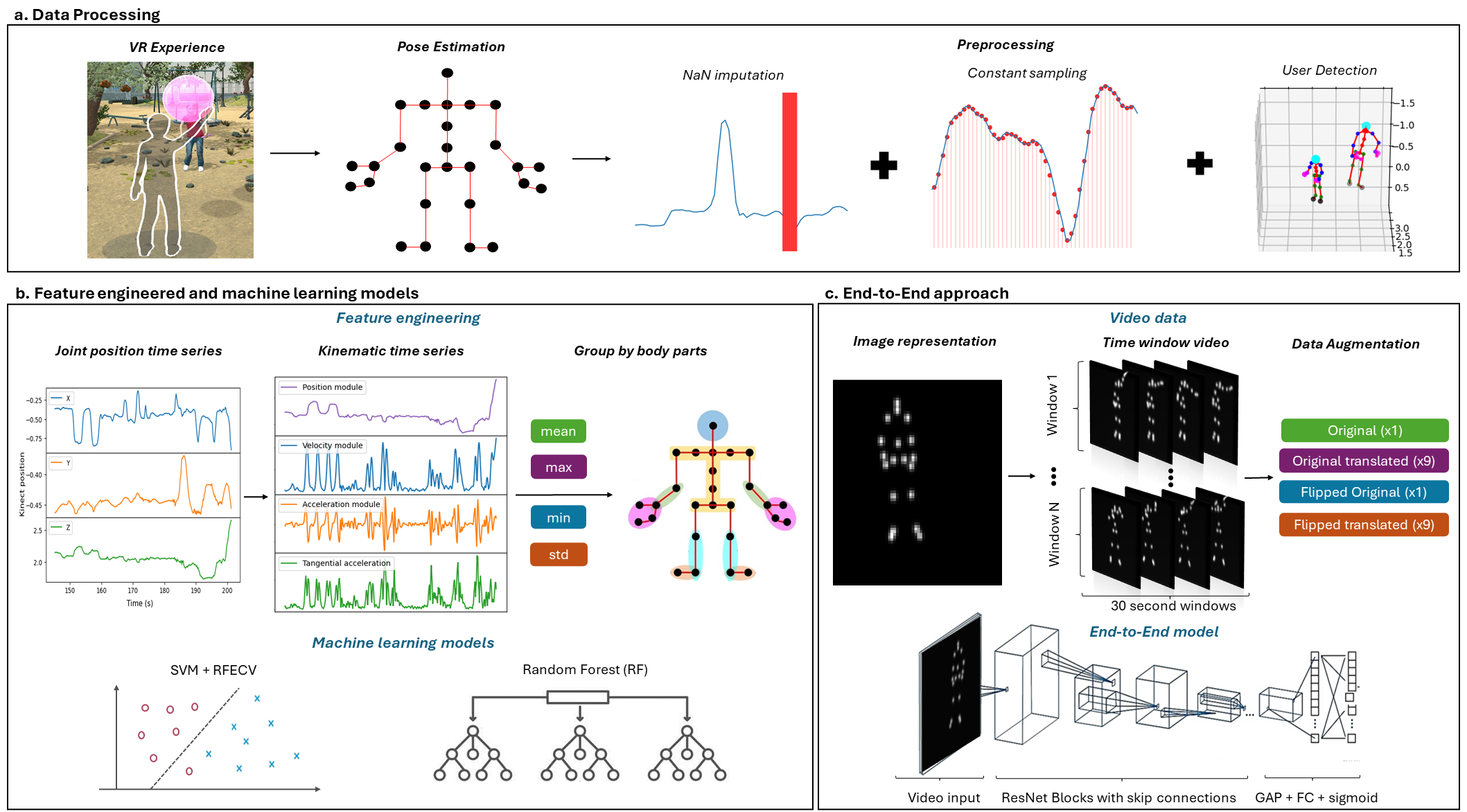}
    \caption{Visual representation of the proposed methodology. Top box represents the processing pipeline, from raw data to the data used for both the feature engineering and end-to-end approaches. Bottom left box represents the feature extraction process and the machine learning models used. Bottom right box represents the video generation process from raw data, the data augmentation process and our proposed end-to-end model.}
    \label{FIG:methods}
\end{figure*}

\subsection{Data Preprocessing}

Initially, the Azure Kinect DK provided the joint positions of the detected bodies in its field of view using a computer vision algorithm in a text log. However, it was found that the text log of joint positions included not only the participants tracked body, but also, in some cases, additional bodies were recognized by the sensor that were present in its field of view, such as the supervising researchers, who were present in the scene. Additionally, sampling rate was not uniform and there were missing values, which could lead to anomalies in engineered features (as it would be the case of instant velocity) or an uneven framerate in the case of generating videos from the tracking data.

To address these issues, we employed two strategies. First, participants were identified throughout the experience using a time-windowed function over time that saved the most repeated tracked body in a time window every step it is applied. Simultaneously, we monitored the centroid of the tracked body to ensure continuous movement. Specifically, to classify a user as valid, two criteria needed to be met: firstly, the centroid of the tracked body had to exhibit continuous movement, defined as a displacement of less than 30cm between frames. Secondly, the user selected should be the most frequently tracked body within 10-second windows. This approach allowed us to extract tracking data exclusively for the participants in the experiment, provided that they were the most commonly tracked bodies and displayed smooth movement, as per our two primary assumptions. This cleaning process was validated visually for all users, and it proved to eliminate all tracked body duplicates. Then, a forced sampling rate of 10Hz was applied to all Azure Kinect DK text files by interpolating the data and eliminating the rest of the datapoints in order to solve the uneven framerate. This method also solved the appearance of missing values, which were interpolated throughout the scene using the position of each joint before and after the missing value.

\subsection{Hand-crafted features approach}

\subsubsection{Data processing for feature engineering}

In order to obtain the hand-crafted features inspired by previous literature \cite{autism-crippa, autism-vabalas, autism-forti, autism-alcaniz, autism-simeoli, autask-zhao}, such as velocity, position, or acceleration, we implemented a set of mathematical functions. These functions included the computation of the Euclidean distance between joints in consecutive frames, the calculation of derivatives for the time series of each joint, and the determination of the magnitude of vectors based on the three spatial coordinates.

Using these mathematical functions, we obtained time series data for various physical parameters, including displacement, velocity, acceleration, and tangential acceleration. Subsequently, we extracted relevant features from these time series. Specifically, we computed key statistical descriptors: the mean, variance, maximum, and minimum values, for the time series data of each joint. However, this approach resulted in a substantial number of kinematic features that exhibited high correlation with each other due to joint proximity. To address this issue, we performed an aggregation step, wherein neighboring joint features were combined into larger body parts. This aggregation was achieved by computing the mean of the corresponding features for each independent body group. The defined body part groups included head, body and left and right arms, legs, feet, and hands, respectively (see \ref{FIG:methods}).


\subsubsection{Feature engineered machine learning models}
\label{ml-models}

Once the hand-crafted features were generated for every subject and task, initial classifiers were explored using some of the training partitions. As a result, it was noted that non-linear machine learning models, such as rbf-SVM or kPCA with classifiers, exhibited comparable performance compared to simpler linear algorithms. Therefore, we have incorporated the linear classifiers to enhance interpretability and reduce computational complexity. Specifically, we have employed SVM with Recursive Feature Elimination(RFECV) and Random Forest classifiers, which are widely used in previous literature \cite{autism-crippa, autask-zhao, autism-vabalas, autism-alcaniz}.

Our model fine-tuning aimed to maximize the Receiver Operating Characteristic - Area Under the Curve (ROC-AUC), albeit through slightly distinct approaches. Specifically, for the LinearSVM with RFECV, we executed a three-tiered nested cross-validation process, delineated as follows:
\begin{enumerate}
\item Feature Selection using RFECV: This entailed the selection of the optimal number of features through stratified k-fold cross-validation, with five folds considered for each C regularization parameter, ranging from $2^{-6}$ to $2^7$.
\item	Hyperparameter Optimization: Subsequently, having determined the optimal number of features for every C regularization parameter, the most suitable C value was selected. This phase entailed another level of repeated stratified k-fold cross-validation, comprising 5 folds and 6 repetitions, and was conducted on the entire training partition. This step was executed subsequent to the feature selection phase.
\item	Model Assessment: The third and outermost layer of cross-validation was reserved for assessing the overall model performance, as elucidated in Subsection \ref{validation}, under our chosen validation strategy.
\end{enumerate}

In essence, our approach for fine-tuning the LinearSVM with RFECV involved a rigorous three-level nested cross-validation process, wherein the first level focused on feature selection, the second on regularization parameter selection, and the third on model performance assessment.

In contrast, the fine-tuning of the Random Forest Classifier followed a simpler approach. In this case, our grid search efforts concentrated on optimizing the tree depth hyperparameter, spanning the range of ${\mathrm{ma}\mathrm{x} \mathrm{depth}\ }\mathrm{\in }\left[\mathrm{1,2,3,4,5,6}\right]$. Consequently, we engaged in a conventional two-tiered cross-validation process. The initial level, identical to the second level of the LinearSVM, addressed regularization hyperparameter selection, while the subsequent outer level was utilized for model performance evaluation.

\subsection{End-to-end approach}

\subsubsection{Data processing for end-to-end deep learning}

To prepare the input data for our deep learning model, we transformed the preprocessed joint time series into videos, representing them as sequences of pixel intensities. This conversion involved associating each time sample from all joints with an image, effectively generating individual frames that visually represent limb positions. Importantly, the framerate of these videos matched the input's preprocessed joint time series, which was set at 10Hz. This process resulted in video sequences with a resolution of 78 x 64 pixels for each subject and virtual task, an example of which is presented in Figure \ref{FIG:methods}.

To portray body tracking joints in a 2D context, we adopted an algorithm inspired by Haodong \textit{et al.}' \cite{poseconv} approach. Initially, we mapped each pixel in a frame into the Kinect's coordinate system. This mapping was achieved by interpolating distances across a regular grid that spanned between the maximum and minimum $x_j$ and $y_j$ positions of each joint. Equations \ref{EQ:pixelx} and \ref{EQ:pixely} describe this pixel-to-coordinate transformation, where $x_p$ and $y_p$ represent the horizontal and vertical pixel coordinates in the Kinect's space, respectively. Additionally, $h$ and $v$ denote the pixel numbering along the frame's horizontal and vertical axes, while $H$ and $V$ signify the total number of pixels in the frame, horizontally and vertically.

\begin{equation}\label{EQ:pixelx}
    x_p(h)=\mathrm{min} \left(x_j\right)\ +\frac{h}{H}\left(\mathrm{max} \left(x_j\right)-\mathrm{min} \left(x_j\right)\right)
\end{equation}
\begin{equation}\label{EQ:pixely}
    y_p(v)=\mathrm{min} \left(y_j\right)\ +\frac{v}{V}\left(\mathrm{max} \left(y_j\right)-\mathrm{min} \left(y_j\right)\right)
\end{equation}

Subsequently, pixel intensities for each frame were calculated based on the proximity of each pixel to the joints. Specifically, we employed the $x$ and $y$ components of the joints as the means for 2D normal distributions. The pixel intensity was then computed as the cumulative probability of a pixel being sampled from these Gaussian distributions, as outlined in Equation \ref{EQ:gaussian}. Here, $I\left(x_p(h),y_p(v)\right)$ represents the pixel intensity at position $h,v$, $\sigma$ denotes the standard deviation parameter for the normal distribution, and $x_j$ and $y_j$ indicate the positional components of each joint. Importantly, this representation heightened the intensities of pixels closer to the joints, effectively highlighting the joints positions in the frame, while background pixels received intensities close to zero. 

\begin{equation}\label{EQ:gaussian}
    I\left(x_p(h),y_p(v)\right)=\ \sum_{j\ \epsilon \ Joints}{\frac{1}{\sigma \sqrt{2\pi }}e^{{-\frac{1}{2}\left(\frac{\left(x_p,y_p\right)-\ \left(x_j,y_j\right)}{\sigma }\right)}^2}}
\end{equation}

\subsubsection{Data augmentation for Deep Learning}

During the video generation process, it became apparent that participants were not in identicals horizontal starting positions, leading to minor horizontal discrepancies across subjects. These discrepancies had the potential to divert the model's focus towards distinguishing subjects based on position rather than capturing general movement characteristics. Additionally, due to the intricate nature of deep learning models, a substantial volume of examples is required for effective generalization.

To address these concerns and enhance model generalization, we employed data augmentation techniques. Specifically, for every video and user, we generated an additional set of 10 videos in both training and testing partition. This generation involved introducing a random horizontal variation to all joint positions. The extent of variation was determined by adding a constant random value ($\varepsilon$), sampled from a normal distribution characterized by a mean ($\mu_x$) of 0 and a standard deviation ($\sigma_x$) of 0.35. The choice of $\sigma_x$ was deliberate, ensuring that approximately 99\% of the samples fell within the range of $x_j \in [\mathrm{min}(x_j), \mathrm{max}(x_j)]$. Furthermore, we created another set of 10 videos by horizontally flipping each of the previously generated ones. This process resulted in a total of 20 videos derived from each original video sample, effectively augmenting the dataset size and introducing valuable variability.

\subsubsection{End-to-end Deep Learning model}

\begin{table}
    \fontsize{10.4pt}{10.4pt}\selectfont
    {\renewcommand{\arraystretch}{1}
    \begin{tabular}{cc} 
    \hline 
    $\boldsymbol{\mathrm{Stage}}$\textbf{} & $\boldsymbol{\mathrm{PoseConv3D}}\boldsymbol{\mathrm{\ (}}\boldsymbol{\mathrm{SlowOnly}}\boldsymbol{\mathrm{)}}$\textbf{} \\ \hline 
    $\boldsymbol{\mathrm{Data}}\boldsymbol{\mathrm{\ }}\boldsymbol{\mathrm{Layer}}$\textbf{} & $\mathrm{Uniform:T\times (}\mathrm{78\times 64),\ 1}$ \\ \hline 
    $\boldsymbol{\mathrm{Stem}}\boldsymbol{\mathrm{\ }}\boldsymbol{\mathrm{Layer}}$\textbf{} & $[1\times 7^2,\mathrm{\ 32}]\times 1$ \\ \hline 
    $\boldsymbol{\mathrm{Stage}}\boldsymbol{\mathrm{\ }}\boldsymbol{\mathrm{1}}$\textbf{} & $\left[ \begin{array}{c}
    1\times 1^2,\ 32 \\ 
    1\times 3^2,\ 32 \\ 
    1\times 1^2,\ 128 \end{array}
    \right]\times 4$ \\ \hline 
    $\boldsymbol{\mathrm{Stage}}\boldsymbol{\mathrm{\ }}\boldsymbol{\mathrm{2}}$\textbf{} & $\left[ \begin{array}{c}
    3\times 1^2,\ 64 \\ 
    1\times 3^2,\ 64 \\ 
    1\times 1^2,\ 256 \end{array}
    \right]\times 6$ \\ \hline 
    $\boldsymbol{\mathrm{Stage}}\boldsymbol{\mathrm{\ }}\boldsymbol{\mathrm{3}}$\textbf{} & $\left[ \begin{array}{c}
    3\times 1^2,\ 128 \\ 
    1\times 3^2,\ 128 \\ 
    1\times 1^2,\ 512 \end{array}
    \right]\times 3$ \\ \hline 
    & $\mathrm{Global\ Average\ Pooling\ (GAP)}$ \\
    $\boldsymbol{\mathrm{Output}}\boldsymbol{\mathrm{\ }}\boldsymbol{\mathrm{Stage}}$\textbf{}& $\mathrm{Fully\ connected\ Layer}$ \\
    & $(\mathrm{FC})\mathrm{Sigmoid\ activation}$ \\ \hline 
    \end{tabular}} \quad
    \caption{End-to-End ResNet 3DCNN architecture.} 
    
    \label{TB:deeplmodel}
\end{table}

In our pursuit of ASD classification using deep learning, we trained and implemented from scratch the PoseConv3D architecture, a tailored spatio-temporal residual 3DCNN model originally introduced by Haodong \textit{et al.} \cite{poseconv} for action recognition tasks in 15-second body tracking videos. This architecture showcases potential for ASD detection through movement data, harnessing its capacity to capture spatial and temporal details within movement patterns. Drawing inspiration from Haodong \textit{et al.}' work, our model operates on videos featuring body-tracked joints superimposed on a consistent background. This approach empowers the model to focus on pertinent body parts, effectively filtering out background noise and RGB video complexities. Moreover, the spatio-temporal nature of the network, with convolutions spanning both spatial and temporal dimensions, enables the identification of crucial motion patterns vital for classifying children's movements within the virtual environment.

In terms of architectural details, our proposed deep neural network closely follows the PoseConv3D SlowOnly model introduced by Haodong \textit{et al.} \cite{poseconv} for action recognition tasks, maintaining consistent layer counts, pooling layers, and kernel sizes. However, we introduced a modification to the final activation layer to tailor it to our specific binary classification problem of differentiating between ASD and non-ASD cases. In contrast to Haodong \textit{et al.}' original model, we employed a single-neuron sigmoid activation for this layer. Our modification of PoseConv3D architecture is illustrated in Table \ref{TB:deeplmodel}, where the dimensions of kernels are denoted by $T$ × $S^{2}, C$ for temporal, spatial and channel sizes and GAP denotes global average pooling. Moreover, certain hyperparameters were determined based on unique considerations specific to our study. Computational and memory constraints guided our choices for batch size and temporal sample size. Consequently, we selected a batch size of 3 and designed our subject samples to span 30 seconds. This decision ensured efficient memory utilization while accommodating the need for capturing a broader temporal context in ASD classification. Unlike action recognition tasks, which may rely on shorter video segments, ASD classification benefits from more extensive temporal information. This extension of video length from Haodong \textit{et al.}' original 15 seconds to 30 seconds was essential for providing our model with the necessary context for accurate predictions. It's worth noting that the tasks within the virtual experience typically lasted 1 to 3 minutes, resulting in multiple videos for each user, each covering a 30-second interval with 15-second overlaps. Subsequently, the final prediction for each user was generated by aggregating all voting predictions derived from all the overlapping windowed videos.

\begin{figure*}
    \centering
    \includegraphics[width = 170mm, height = 72mm]{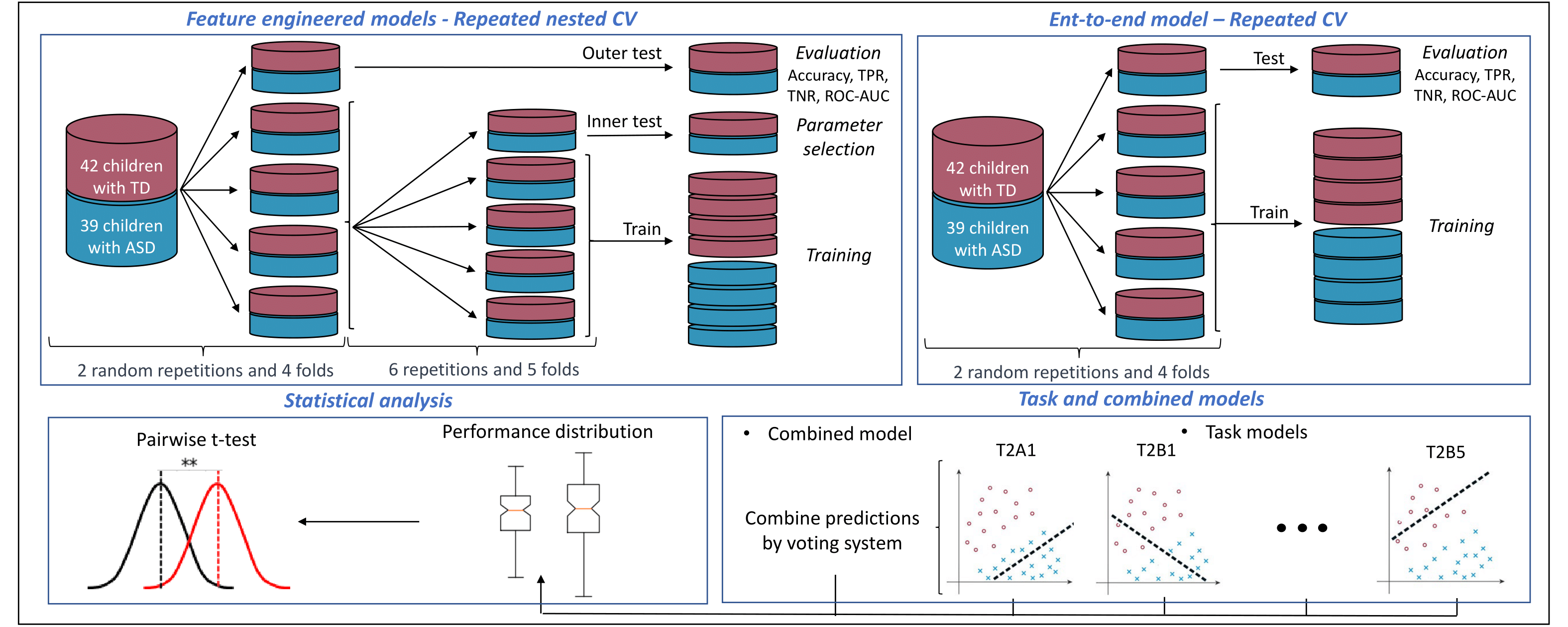} 
    \caption{Representation of the validation strategy. Top boxes show our cross validation strategies for both the end-to-end and feature engineered models. Bottom right represents our voting system for the ensemble model, which combines task-specific model predictions. Bottom left depicts the pairwise statistical analysis used for model comparison.}
    \label{FIG:validation}
\end{figure*}

Conversely, the establishment of suitable values for the number of epochs and the learning rate relied on initial exploratory experiments conducted using a single training set. Following this preliminary validation, it was determined that a training comprising 200 epochs, produced favorable outcomes without introducing the risk of overfitting. To optimize our model, we employed a cross-entropy loss function in conjunction with Stochastic Gradient Descent (SGD) with a learning rate set to 0.01. A \textit{ReduceOnPlateau} strategy was applied, incorporating a patience parameter of 10 epochs to dynamically adjust the learning rate as training progressed. Additionally, given the limited validation data and the potential overfitting, we implemented an early stopping mechanism with a patience of 25 epochs to provide protection against this risk.

\subsection{Task-Specific and voting models}

The virtual experience offers a range of diverse tasks, each designed to engage users differently. To thoroughly evaluate the efficacy of both deep learning and feature engineering approaches in ASD classification, we constructed distinct models tailored to each specific task within the virtual environment. This approach allowed us to create individual models for each task, enabling a thorough comparison between feature engineering techniques and the deep learning model in various contexts.

Furthermore, we introduced an ensemble method to consolidate predictions generated by these task-specific models. Instead of combining all predictions into one, we established three distinct ensemble models: one for the task-specific LinearSVM models, one for the task-specific Random Forest models, and one for the task-specific end-to-end model. For each ensemble model, the predictions from task-specific models are aggregated by calculating the mean of the predicted probabilities. By aggregating predictions from feature-engineered models within each task category and, similarly, for the end-to-end model, we can evaluate their collective performance in ASD classification, enabling a more meaningful model comparison.

\subsection{Validation Strategy}
\label{validation}

Our validation strategy for assessing model performance involves a subject-dependent repeated stratified k-fold approach with 4 folds and 2 repetitions, totaling 8 folds (refer to Figure \ref{FIG:validation}). Specifically, we partition participants into stratified folds based on their respective groups, ensuring a proportional representation from each group within each fold. Following this division into training and testing partitions, we create task-specific datasets. For each of the 12 virtual tasks, we establish two datasets: one containing task-specific hand-crafted features and another with task-specific kinematic windowed videos, resulting in a total of 24 datasets.

Within each fold, we train all models, including feature-engineered and deep learning models, using the generated datasets. Deep learning models utilize the 12 video datasets, while machine learning models (Random Forest and LinearSVM) operate on the 12 feature-engineered datasets. Feature-engineered models undergo additional fine-tuning during training (see subsection \ref{ml-models}), making our strategy a nested cross-validation. In contrast, end-to-end models use fixed hyperparameters and do not require fine-tuning. Following training, all models are tested on the corresponding 12 test datasets, which share the same subjects to ensure a fair model comparison. After training, we test all models on corresponding test datasets with the same subjects to ensure a fair comparison. We aggregate model predictions using a voting system to evaluate the performance of feature-engineered and end-to-end models across all tasks and their ensemble performance.

In order to validate each model’s performance a set of metrics were considered: accuracy (i.e., percentage of subjects correctly recognized), true positive rate (i.e., percentage of ASD subjects correctly labelled), true negative rate (i.e., percentage of control subjects recognized as control), and Receiver Operating Characteristic Area Under the Curve (ROC-AUC), which describes the model's ability to differentiate between positive and negative classes, with a value of 0.5 indicating performance equivalent to random classification and a value of 1 signifying perfect discrimination.

\section{Results}

\begin{table*}   
    \centering
    \fontsize{9.7pt}{9.7pt}\selectfont
    \begin{NiceTabular}{|c|c|c|c|c|c|}[vlines]
    
    \hline 
        Game & Model & Accuracy & TPR & TNR & AUC  \\  \hline
        \Block{3-1}{EF} & PoseConv3D & 72±08 & 72±17 & 74±12 & \textbf{78±08}  \\ 
        ~ & SVM+RFECV & \textbf{73±08} & 58±22 & \textbf{85±07} & 72±15  \\ 
        ~ & Random Forest & 72±08 & 57±28 & 84±14 & 75±17  \\ \hline
        \Block{3-1}{I2} & PoseConv3D & 56±16 & \textbf{80±18} & 38±35 & \textbf{71±11}  \\ 
        ~ & SVM+RFECV & 56±20 & 50±32 & \textbf{65±30} & 63±28  \\ 
        ~ & Random Forest & \textbf{57±16} & 53±19 & 60±24 & 52±18  \\ \hline
        \Block{3-1}{PEAP} & PoseConv3D & \textbf{71±09} & \textbf{87±18} & 59±20 & \textbf{82±10}  \\ 
        ~ & SVM+RFECV & 59±14 & 62±15 & 57±16 & 63±18  \\ 
        ~ & Random Forest & 64±14 & 64±19 & \textbf{64±15} & 64±16  \\\hline
        \Block{3-1}{T2A1} & PoseConv3D & 83±12 & 75±19 & \textbf{88±07} & 84±11  \\ 
        ~ & SVM+RFECV & \textbf{85±03} & \textbf{85±13} & 85±13 & \textbf{90±06}  \\ 
        ~ & Random Forest & 84±07 & 81±12 & 86±08 & 87±10  \\ \hline
        \Block{3-1}{T2A2} & PoseConv3D & \textbf{81±11} & \textbf{74±15} & 85±12 & \textbf{79±15}  \\ 
        ~ & SVM+RFECV & 65±06 & 54±16 & 72±11 & 64±19  \\ 
        ~ & Random Forest & 75±12 & 50±20 & \textbf{90±13} & 77±14  \\ \hline
        \Block{3-1}{T2A3} & PoseConv3D & 78±13 & 68±17 & 86±12 & \textbf{86±12}  \\ 
        ~ & SVM+RFECV & 79±11 & 73±19 & 84±14 & 84±11  \\ 
        ~ & Random Forest & \textbf{81±12} & \textbf{75±21} & \textbf{86±09} & 86±14  \\ \hline
        \Block{3-1}{T2A4} & PoseConv3D & \textbf{69±15} & 47±22 & \textbf{83±16} & \textbf{81±15}  \\ 
        ~ & SVM+RFECV & 58±23 & 35±30 & 73±27 & 62±26  \\ 
        ~ & Random Forest & 66±18 & \textbf{50±28} & 77±18 & 69±24  \\ \hline
        \Block{3-1}{T2A5} & PoseConv3D & \textbf{75±09} & \textbf{51±25} & \textbf{89±10} & \textbf{81±10}  \\ 
        ~ & SVM+RFECV & 66±15 & 38±31 & 86±9 & 65±27  \\ 
        ~ & Random Forest & 64±14 & 35±33 & 80±11 & 71±21  \\ \hline
        \Block{3-1}{T2B1} & PoseConv3D & 78±11 & 63±19 & \textbf{89±12} & \textbf{89±06} \\ 
        ~ & SVM+RFECV & \textbf{82±08} & \textbf{78±13} & 85±15 & 88±08  \\ 
        ~ & Random Forest & 76±09 & 61±23 & 86±12 & 85±09  \\ \hline
        \Block{3-1}{T2B2} & PoseConv3D & 76±10 & 54±17 & \textbf{91±10} & 72±12  \\ 
        ~ & SVM+RFECV & \textbf{77±12} & \textbf{70±17} & 82±14 & 81±09  \\ 
        ~ & Random Forest & 74±11 & 61±16 & 83±13 & \textbf{82±08}  \\ \hline
        \Block{3-1}{T2B3} & PoseConv3D & 73±09 & \textbf{66±18} & 78±12 & \textbf{83±09}  \\ 
        ~ & SVM+RFECV & 76±12 & 65±17 & 82±16 & 82±10  \\ 
        ~ & Random Forest & \textbf{79±11} & 55±15 & \textbf{96±08} & 76±09  \\ \hline
        \Block{3-1}{T2B4} & PoseConv3D & \textbf{74±09} & \textbf{55±14} & 87±11 & 77±11  \\ 
        ~ & SVM+RFECV & 67±08 & 46±06 & 84±12 & 75±07  \\ 
        ~ & Random Forest & 69±07 & 46±16 & \textbf{87±09} & \textbf{78±06}  \\ \hline
        \Block{3-1}{Mean} & PoseConv3D & \textbf{74±03} & \textbf{66±03} & 79±07 & \textbf{80±03}  \\ 
        ~ & SVM+RFECV & 70±06 & 60±08 & 78±07 & 74±08  \\ 
        ~ & Random Forest & 72±04 & 57±06 & \textbf{82±05} & 75±06  \\ \hline
        \Block{3-1}{Global Voting} & PoseConv3D & 77±10 & \textbf{74±20} & 80±07 & \textbf{86±10}  \\ 
        ~ & SVM+RFECV & 77±13 & 64±20 & 87±17 & 82±13  \\ 
        ~ & Random Forest & \textbf{80±11} & 68±19 & \textbf{90±10} & 83±16  \\ \hline
    \end{NiceTabular}
    \caption{Outer test performance results for end-to-end and handcrafted feature models, ensemble voting, and mean performance aggregates \label{table:performance}}
\end{table*}

\begin{figure*}
    \centering
    \includegraphics[width = 160mm, height = 55mm]{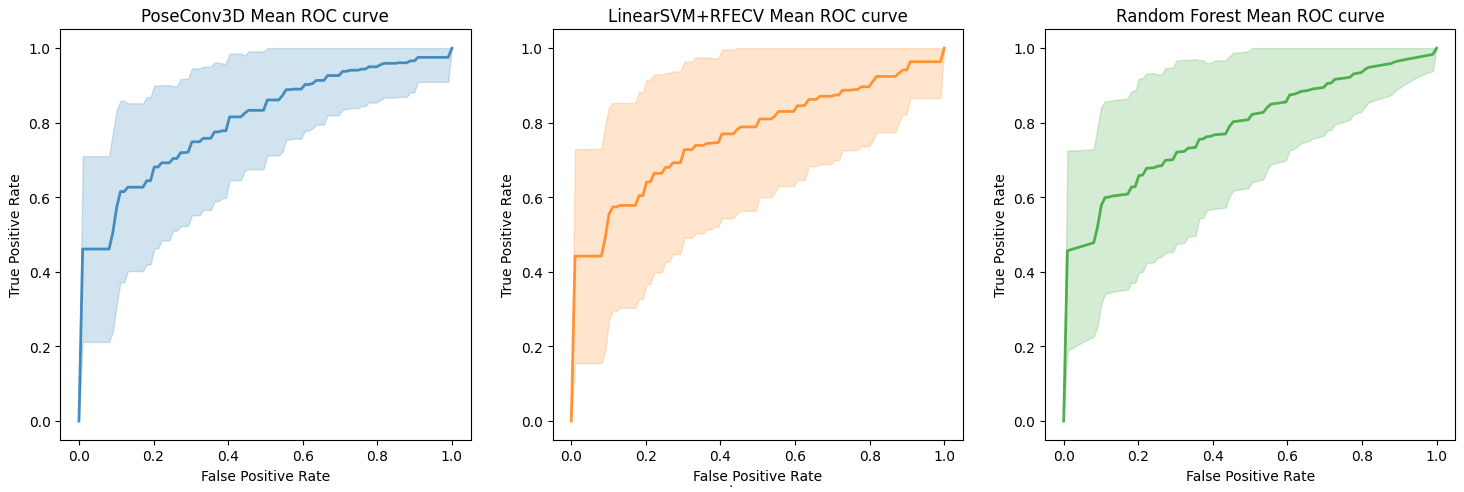}
    \caption{Mean ROC curves and standard deviations (highlighted) across models for all folds and games.}
    \label{FIG:ROC_AUCs}
\end{figure*}

\begin{figure}
    \centering
    \includegraphics[width = 60mm, height = 60mm]{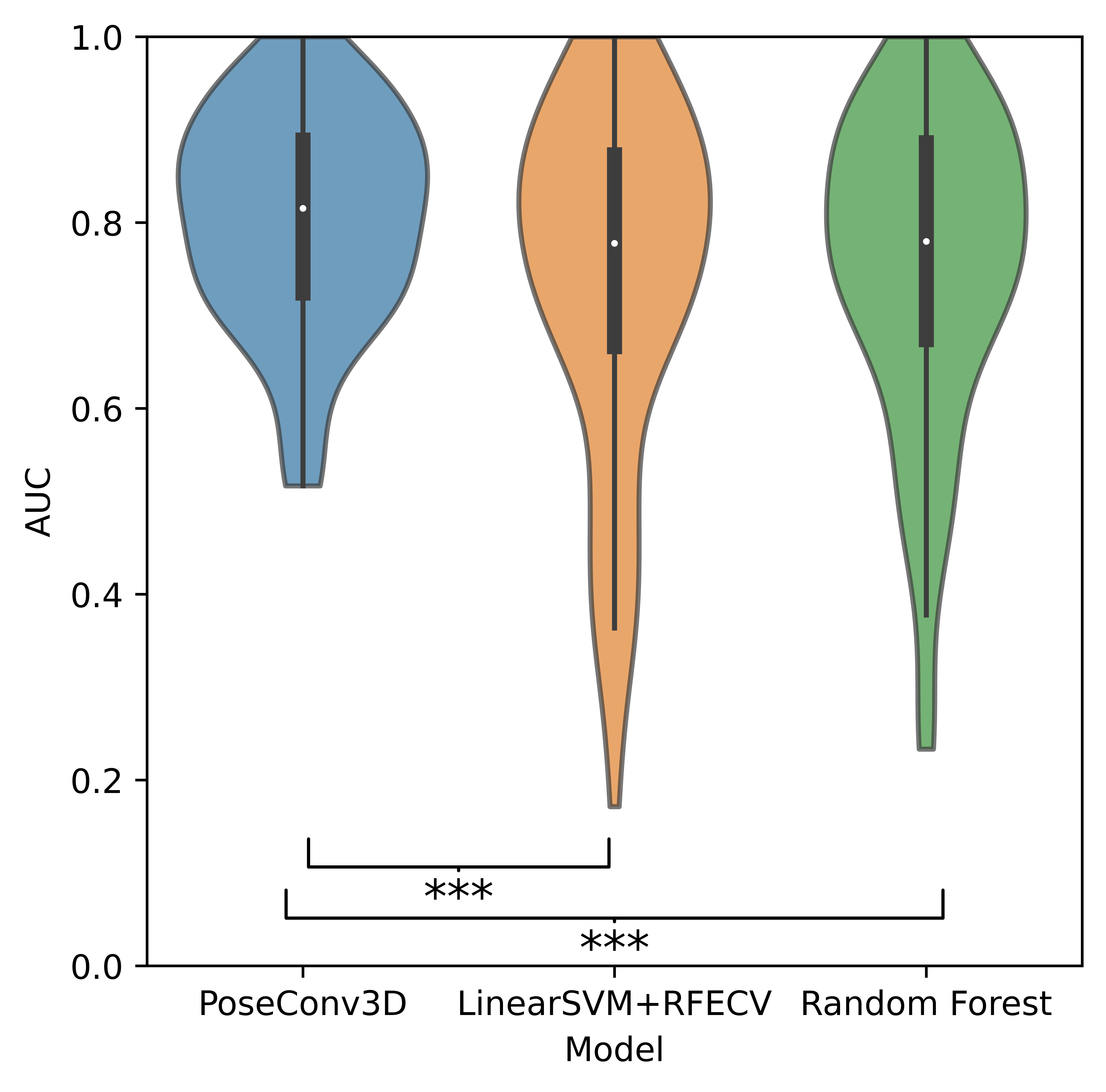}
    \caption{Levene differences across models independently of the game \label{FIG:levene}}
\end{figure}

Table \ref{table:performance} presents the mean and variance of the test performance results across all folds and for each of the proposed models: the end-to-end models and the models trained with handcrafted features for each game, along with their ensemble global voting and the aggregated mean performance across all games. Variability was observed across games and k-fold splits, with some tasks yielding higher performance than others. Notably, tasks related to gross motor coordination (e.g., T2A1, T2A3, T2B1) achieved high accuracies of over 80\% and ROC-AUC values of 0.85 to 0.90, denoting outstanding performance for binary classification.

Regarding the feature-engineered models (SVM+RFECV and Random Forest), they outperformed the end-to-end model in 7 out of 12 goal-oriented models in terms of accuracy. Specifically, the LinearSVM surpassed the deep learning approach in 6 out of 12 games, while the Random Forest surpassed it in 4 out of 12. Although the feature-engineered models generally showed slightly better accuracies, their largest improvement over deep learning was only 6\% in the T2B3 task, with the rest showing smaller improvements. On the other hand, deep learning outperformed the feature-engineered models by up to 9\% in T2A5, 7\% in PEAP, and 6\% in T2A2. These differences are further evident in the mean accuracies across games, favoring the deep learning model. It achieved a higher mean accuracy despite outperforming feature-engineered models in only 5 out of 12 tasks. However, when deep learning did outperform the feature-engineered models, it did so more significantly than when the feature-engineered models outperformed the deep learning model. 

Regarding ROC-AUC, the end-to-end deep learning model showcased consistently higher results, outperforming feature-engineered models in 9 out of 12 games. The LinearSVM + RFECV surpassed the ROC-AUC in 1 game, while Random Forest did so in 2. The end-to-end model showed significant improvements in ROC-AUC compared to feature-engineered models, with some cases showing over 0.10 improvements, particularly in PEAP, where it achieved a remarkable 0.18 increase. These results support the findings from the accuracy results, indicating that the end-to-end model significantly improved performance in games where feature-engineered models performed poorly. 

Nevertheless, these differences in accuracy and ROC-AUC between end-to-end and feature-engineered models can be partially explained by the TPR and TNR metrics. Feature-engineered models consistently exhibited high TNR and lower TPR, while the end-to-end model achieved a more balanced TPR and TNR, resulting in higher ROC-AUC values. Although feature-engineered models achieved slightly higher accuracies more frequently, particularly due to the dataset's slight negative class bias, the end-to-end model demonstrated better overall balance and consistency in distinguishing between classes.

\subsection{Model Comparison and Statistical Analysis}

In this study we conducted a comprehensive comparison between feature-engineered and end-to-end models across various application contexts, assessing their generalization capabilities in diverse kinematic tasks. Figure \ref{FIG:ROC_AUCs} represents the ROC curves of each model across all folds and tasks. The highlighted area for each model represents the standard deviation of the mean in each point of the curve.  To facilitate model comparison across tasks, we aggregated the ROC-AUC results from all folds and tasks into performance distributions. Figure \ref{FIG:levene} provides a visual representation of the ROC-AUC scores for both feature-engineered models and the deep learning model across all aggregated folds and tasks.

Statistical analysis, including a T-test, was performed to compare all models, revealing no statistically significant differences in their mean ROC-AUC distributions (p $>$ 0.05). However, the application of a Levene test indicated significant differences in variances among the AUC of the models. Instances marked with an asterisk (*) in Figure \ref{FIG:levene} signify statistical significance, highlighting situations where performance variance differences between models are significant. Figure \ref{FIG:levene} illustrates that while differences in performance are not highly pronounced, the end-to-end deep learning model exhibits lower model variance and a more stable distribution. This is also visually supported by Figure \ref{FIG:ROC_AUCs}, where it can be appreaciated that the standard deviation is lower for the end-to-end deep learning model throughout the ROC curve.

\section{Discussion}

In this study, our primary objective is to assess the performance of the proposed end-to-end deep learning model while comparing it to hand-crafted feature models across various scenarios. To achieve this objective, we recruited a total of eighty-one children aged between 3 and 7 years, segregating them into two groups: ASD, comprising children with a confirmed diagnosis of the disorder, and a control group. Participants engaged in a VR scenario, where they interacted with a virtual environment and completed diverse tasks. An RGB-D camera recorded their body movements during these tasks, serving as input for training an end-to-end deep learning model based on spatio-temporal kinetic data, as well as two models trained with hand-crafted features characterizing movement.

\subsection{Model Performance and Statistical Differences}

Our findings indicate that feature-engineered models exhibited higher accuracy than the end-to-end model across certain tasks, such as tasks involving touching moving objects or action imitation. However, these improvements were typically modest, ranging from 1\% to 6\%, with an average improvement of 2.6\%. Conversely, there were instances where the end-to-end model outperformed the hand-crafted models, with accuracy improvements ranging from 3\% to 7\% and averaging 6\%, leading to the end-to-end model achieving a superior mean accuracy across tasks. Another important outcome is that our end-to-end model showed a higher mean TPR than the feature-engineered models, with only a slight decrease in TNR. This resulted in a more balanced TPR-to-TNR ratio, enhancing class distinguishability, particularly evident in the ROC-AUC. On average, the end-to-end model outperformed the feature-engineered models in terms of ROC-AUC by 0.05. It should be noted the superiority of the end-to-end deep learning model was achieved even when placing the model at a comparative disadvantage, as the hand-crafted models underwent fine-tuning using an internal validation strategy for every fold, whereas the end-to-end model's hyperparameters were selected using a single external validation. 

However, we cannot definitively conclude that end-to-end deep learning models consistently outperform feature-engineered models across various tasks and contexts, as our t-test did not reveal significant differences in ROC-AUC mean distributions across all folds and tasks. This could be attributed to greater variability and uncertainty in task-specific performance compared to mean performance differences. Nevertheless, the Levene test indicated significant differences between distributions, suggesting that ROC-AUC exhibited less variability across tasks and folds for the end-to-end model (p $<$ 0.001). This indicates that the end-to-end model is more stable and robust, consistently engineering features that better distinguish both classes across a broader range of contexts and tasks than hand-crafted features. These results emphasize the potential of end-to-end models to adapt across different application contexts. Specifically, results suggest that end-to-end models can effectively extract features even in cases where feature engineering falters, while the opposite isn't as significant.

In summary, although there is no statistically significant evidence confirming that end-to-end models consistently outperform feature-engineered models, they do exhibit statistically higher reliability and consistency in their results across datasets obtained in different contexts in ASD classification. Furthermore, end-to-end models are easier to implement, eliminating the need for defining hand-crafted metrics. However, machine learning models demonstrate higher accuracy in certain tasks, while offering advantages in terms of explainability and ease of training. In essence, end-to-end models provide greater performance stability despite their increase in complexity and decrease in interpretability when compared to hand-crafted machine learning models.

\subsection{Comparison with State-Of-the-Art}

In the realm of ASD classification, few studies have explored full-body tracking. However, these studies often grapple with validation limitations \cite{autism-crippa, autism-simeo, autask-zhao}. To ensure the practical applicability of ASD classification, it is paramount to accurately identify the disorder. Researchers should focus on subject-dependent cross-validations and meticulous separation of unseen test data. Examples of practices to be avoided include training models with various feature sets or hyperparameters sets and reporting the best result. These examples all fall under the umbrella of model selection strategies, which must be externally validated with real unseen test datasets to assess their effectiveness. This work distinguishes itself by emphasizing model robustness and reliability, investigating model performance and its standard deviation across folds. Consequently, our study stands as one of the first to extensively validate its findings and report the standard deviation of the model test performance, which is advisable given the size the datasets in these studies. To date, the only previous study that prioritized validation in the ASD classification domain is the work of Vabalas \textit{et al.} \cite{autism-vabalas}, which achieved lower accuracy (73\%) and reported greater model variability across folds.

Another contribution to the existing literature is our deep learning 3DCNN ResNet strategy, which outperforms the current state-of-the-art in terms of accuracy. The predominant end-to-end deep learning ASD classification literature is primarily led by Kojovic \textit{et al.} \cite{autism-kojovic} and Zunino \textit{et al.} \cite{autask-zunino}, who reported accuracies of 80.9\% and 75\%, respectively. Although our ensemble deep learning model attains a slightly lower accuracy than that of Kojovic \textit{et al.} \cite{autism-kojovic}, with our model achieving 77\% (SD = 10\%), our best task-specific deep learning model reaches an accuracy of 85\% (SD=3\%) with a 1-to-3-minute sample, surpassing their results. It is worth noting that Kojovic \textit{et al.} \cite{autism-kojovic} utilized much longer 1-hour video samples to achieve their reported performance. Notably, Kojovic \textit{et al.} \cite{autism-kojovic} reported that using shorter 10-minute video segments for training reduced their accuracy to approximately 70\%, and it dropped even further to around 65\% when using 1-to-5-minute samples. In contrast, Zunino \textit{et al.} \cite{autask-zunino} achieved a lower classification accuracy than our deep learning ensemble model, with a reported accuracy of 75\%. The primary distinction between their works and ours lies in the use of 3D kernels to elaborate features based on both time and space, rather than solely space. In their works, Kojovic \textit{et al.} \cite{autism-kojovic} and Zunino \textit{et al.} \cite{autask-zunino} both extracted spatial features and employed LSTM for temporal classification of the time series. However, our results suggest that enabling time and spatial features to emerge improves performance. This approach presumably enables the network to capture spatial and temporal correlations, crucial for movement classification, resulting in enhanced feature engineering, better performance across various contexts, and ultimately, improved generalization and reliability.

Finally, current literature suggests a variety of tasks for ASD classification using feature-engineered machine learning models. In our work, we concentrated on evaluating the generalizability of these models across various virtual tasks with slightly varying objectives. Alcañiz \textit{et al.} \cite{autism-alcaniz} previously employed multiple tasks in a study involving VR, yet they did not utilize these tasks for model comparison or task validation; their focus was primarily on enhancing classification. This study also contributes to the literature through its comparison of models trained on multiple task-specific datasets, resulting in valuable insights that can be leveraged in future studies, such as design and selection of motor ASD assessment tasks, or feature and model selection. Remarkably, both machine learning and deep learning models concur that, for kinematic models, the most effective task is touching moving objects. However, they diverge when it comes to tasks related to picking up and dropping virtual objects, where current feature-engineered models underperform, while end-to-end deep learning models excel.

\subsection{Future Work and Limitations}

While our study has provided insights into ASD motor assessment models through the newly adapted end-to-end model and its comparison to feature-engineered models, several limitations and avenues for future research merit consideration.

Firstly, the choice of the Azure Kinect for data capture, while providing reportedly moderate to excellent tracking accuracy for clinical movement tasks \cite{doi:10.1371/journal.pone.0279697}, may lack the precision required for finer motor tasks such as handwriting or precise object manipulation. Advancements in tracking technology capable of high-precision, full-body tracking would be essential to broaden the scope of ASD motor research and explore a wider range of motor difficulties with greater accuracy.

Additionally, our methodological approach, focusing on training individual models per motor task, may benefit from the development of a more generalized model capable of predicting ASD-related movement patterns across tasks. Such a model would necessitate a larger and more diverse dataset, potentially sourced from videos capturing ASD individuals in naturalistic settings, to capture the full spectrum of ASD motor behaviors.

Moreover, while specific motor tasks may effectively classify ASD, further investigation is needed to understand the variations in accuracy among different VR tasks. A deeper exploration into the underlying factors contributing to task-specific accuracy variations could provide valuable insights into the nature of ASD motor symptomatology and task difficulty.

Furthermore, exploring alternative model architectures beyond those examined in our study, such as the CNN2D with an LSTM proposed by Kojovic et al.\cite{autism-kojovic}, could offer additional avenues for validating and extending our findings. Additionally, other advanced feature engineering models, such as time series modeling algorithms like ROCKET or leveraging feature extraction from the generated body-tracked video data, hold promise for elucidating the most informative features contributing to ASD classification.

Lastly, expanding the sample size of ASD population represents a critical direction for future research. A larger dataset could facilitate a more reliable validation and further training of the end-to-end model. Additionally, it could enable the development of severity-stratified models and regression analyses based on ADOS severity, leading to more personalized and effective interventions for individuals across the ASD spectrum.

\section{Conclusion}

This study addresses the critical need for more rigorous and dependable ASD classification methods, while simultaneously conducting a thorough comparison between end-to-end deep learning models and feature-engineered counterparts within a virtual reality environment encompassing diverse motor tasks. Our findings indicate that conventional models can indeed achieve state-of-the-art performance, while also providing benefits towards explainability. However, they exhibit less stability and greater variability across different contextual applications within the domain of the study. In contrast, deep learning approaches not only achieve state-of-the-art performance but also showcase remarkable robustness and generalizability, all without necessitating the expertise required for manual feature engineering. In essence, our research highlights that deep learning methods possess the innate ability to autonomously derive meaningful features from movement data, transcending the constraints of specific task contexts and objectives. This inherent adaptability positions deep learning as a potent and reliable tool for ASD classification, shedding light on the intricate movement patterns associated with the disorder.

\section*{Ethics Approval and Consent to Participate}
Approval of ethical and experimental procedures was granted by UPV Ethics Committee under Application No. $P\_06\_04\_06\_20$ and performed in line with the Declaration of Helsinki.

\section*{CRediT authorship contribution statement}
\textbf{Alberto Altozano:}  Writing – original draft, Writing - Review \& Editing, Software, Validation, Formal analysis, Investigation, Data Curation, Visualization. \textbf{Maria Eleonora Minissi:} Conceptualization, Methodology, Validation, Investigation, Writing - Original Draft, Writing - Review \& Editing. \textbf{Mariano Alcañiz:} Conceptualization, Methodology, Supervision, Resources, Project administration, Funding acquisition. \textbf{Javier Marín Morales:} Conceptualization, Methodology, Writing - Review \& Editing, Supervision, Project administration.

\section*{Funding}
This work was supported by the Spanish Ministry of Economy, Industry, and Competitiveness - funded project (IDI - 20201146). It was also co-funded by the European Union through the FEDER of the 
Valencian Community 2014–2020 (IDIFEDER/ 2018/029 and IDIFEDER/2021/038). MEM received partial funding by the Valencian Community (GRISOLIAP/2019/137; CIBEFP/2021/38).

\section*{Declaration of Competing Interests}
The authors have no conflict of interest to declare. This study received neither financial support nor income from any commercial source.

\section*{Declaration of generative AI and AI-assisted technologies in the writing process}
During the preparation of this work the authors used GPT-4 in order to check grammar. After using this tool/service, the authors reviewed and edited the content as needed and take full responsibility for the content of the publication.



\bibliography{bibliografia}
\bibliographystyle{unsrt}
\end{document}